\documentclass[journal,onecolumn]{IEEEconf}

\input epsf
\usepackage{graphicx}
\usepackage{titlesec}
\usepackage{balance} 
\usepackage{stfloats}
\usepackage{newtxtt}
\usepackage[table]{xcolor}
\usepackage{booktabs}
\usepackage{tabularx}
\usepackage{multirow}
\usepackage{listings}
\usepackage{subcaption}
\usepackage{hyperref}
\usepackage{amsmath}
\usepackage{amssymb} 
\usepackage{enumitem}
\usepackage{multicol}
\usepackage{caption}

\hypersetup{colorlinks=true, linkcolor=black,filecolor=black, urlcolor=blue, citecolor=black}

\definecolor{codegreen}{rgb}{0,0.6,0}
\definecolor{codegray}{rgb}{0.5,0.5,0.5}
\definecolor{codepurple}{rgb}{0.58,0,0.82}
\definecolor{codeblue}{rgb}{0.13,0.13,0.82}
\definecolor{backcolour}{rgb}{0.95,0.95,0.92}

\lstdefinelanguage{OpenSCENARIO}[]{Python}{
    morekeywords={use, scenario, map, with, keep, environment, vehicle, stationary_object, var, speed, length, do, serial, parallel, one_of, wait, emit, fall, rise, elapsed},
    alsoletter={@}, 
}

\renewcommand\thesection{\arabic{section}} 

\renewcommand\thesubsectiondis{\thesection.\arabic{subsection}}

\renewcommand\thesubsubsectiondis{\thesubsectiondis.\arabic{subsubsection}}

\begin{document}

\title{\huge\texttt{OSC2Runner}: OpenSCENARIO 2.x Compliant High-Fidelity AV Simulation in CARLA}

\author{Thoshitha Gamage$^{1,*}$, and Lasanthi Gamage$^{2}$\\
	\normalsize $^{1}$Southern Illinois University Edwardsville, Edwardsville, IL, USA\\
	\normalsize $^{2}$Webster University, St. Louis, MO, USA\\
	\normalsize tgamage@siue.edu, lasanthigamage67@webster.edu\\
	\normalsize *corresponding author
}


\maketitle
\begin{abstract}
Scenario-Based Testing predominantly relies on the legacy ASAM OpenSCENARIO 1.x XML standard because existing continuous simulation frameworks lack native execution support for the recently matured v2.x Domain-Specific Language (DSL). Adapting legacy interpreters to evaluate v2.x logic introduces spatiotemporal drift, asynchronous event latencies, and artificial kinematic snapping. Addressing this execution gap, \texttt{OSC2Runner} introduces the first orchestration framework capable of natively mapping the OpenSCENARIO v2.x DSL to CARLA. The framework achieves this by formalizing scenario translation as a compilation pipeline through a multi-pass transpiler architecture. Bypassing static trajectory playback, the architecture synthesizes type-safe Abstract Syntax Trees directly into dynamic deterministic behavior trees (\texttt{py\_trees}) natively mapped to CARLA’s atomic APIs. Empirical validation in highly concurrent adversarial case studies demonstrates tick-by-tick determinism, exact spatial trigger evaluation, and 100.0 ms cross-actor blackboard synchronization. Kinematic analysis proves the strict adherence to continuous environmental boundaries. This architecture transitions Scenario-Based Testing from approximate behavioral interpretation to mathematically rigorous execution, establishing the deterministic backend required for co-simulation, hardware-in-the-loop testing, and automated LLM-driven generation pipelines.
\end{abstract}
\IEEEoverridecommandlockouts
\vspace{1.5ex}

%
\IEEEpeerreviewmaketitle

\section{Introduction}
\label{sec:introduction}

Evaluating Autonomous Vehicles (AVs) requires verification methodologies that scale beyond empirical field testing~\cite{kalra2016driving}. Scenario-Based Testing (SBT) solves this limitation by isolating and executing traffic interactions within virtual environments such as CARLA~\cite{dosovitskiy2017carla}. However, the validity of SBT hinges strictly upon the fidelity of the underlying simulation environment. High-fidelity, deterministic execution becomes especially critical when integrating complex configurations, such as evaluating simulated Electronic Control Units (ECUs) to analyze the causal link where phantom object detection leads to phantom braking. Consequently, any deviation from this fidelity---such as simulation-layer execution latencies or inaccurate asynchronous event handling---fundamentally invalidates the intended test parameters.

Standardizing scenario definitions ensures test reproducibility across platforms. Early standards, such as ASAM OpenSCENARIO XML 1.x~\cite{asam2020openscenarioxml}, optimized predictable trajectory playback but lacked the expressiveness required for complex scenario logic. To resolve this, ASAM finalized the OpenSCENARIO Domain-Specific Language (DSL) v2.1 in 2024~\cite{asam2024openscenariodsl}. The declarative DSL separates abstract, logical, and concrete layers. It replaces rigid parameterization with native variables, semantic validation, and composable behaviors, reducing script length and complexity. Table~\ref{tab:osc_comparison_1_3} contrasts key structural and operational differences between the XML and DSL schemas.

\begin{table}[!hbt]
\centering
\small
\rowcolors{2}{gray!10}{white}
\caption{Comparison of OpenSCENARIO v1.3 (XML) and v2.1 (DSL)}
\label{tab:osc_comparison_1_3}
\begin{tabularx}{\columnwidth}{@{} 
    >{\raggedright\arraybackslash}p{0.22\columnwidth} 
    >{\raggedright\arraybackslash\hsize=0.85\hsize}X 
    >{\raggedright\arraybackslash\hsize=1.15\hsize}X 
    @{}}
\toprule
\rowcolor{white} \textbf{Feature} & \textbf{OSC 1.3 (XML)} & \textbf{OSC 2.1 (DSL)} \\
\midrule
\textbf{Format} & Schema-based (XML) & Text-based DSL (EBNF) \\
\textbf{Hierarchy} & Storyboard  $\rightarrow$ Act  $\rightarrow$ Maneuver  $\rightarrow$ Event  $\rightarrow$ Action & Declarative, intent-driven structure \\
\textbf{Logic} & Trigger-based conditions & Full expressions, conditions, and logic \\
\textbf{Param\-eter\-ization} & Parameters and catalogs & Native variables and reusable abstractions \\
\textbf{Modularity} & Limited reuse via catalogs & High modularity and composability \\
\textbf{Behavior Modeling} & Explicit action definitions & Actions + Modifiers + Conditions separation \\
\textbf{Semantic Validation} & Syntax (XSD) only & Supports ontology-based semantic validation \\
\textbf{Readability} & Verbose ($\sim$150--300 lines) & Concise ($\sim$20--50 lines) \\
\textbf{Integration} & Widely supported in tools & Requires DSL parser/compiler, limited support \\
\textbf{Execution Model} & Event-driven storyboard execution & Intent-driven, composable execution model \\
\textbf{Extensibility} & Limited & High (DSL + custom constructs) \\
\bottomrule
\end{tabularx}
\end{table}

Integrating the v2.x DSL into continuous simulators like CARLA presents an implementation gap. Existing interpreter-based orchestration frameworks, such as \texttt{ScenarioRunner}~\cite{carlateam2024scenariorunner}, predominantly support the legacy XML standard, offering compatibility restricted only to early, deprecated DSL drafts. Consequently, these legacy tools cannot parse the finalized, bifurcated v2.x standard libraries (\texttt{types.osc} and \texttt{domain.osc}) and fail to maintain deterministic execution loops for concurrent actions. The resulting execution drift prevents researchers from leveraging the v2.x standard for high-fidelity validation.

Addressing this gap, \texttt{OSC2Runner} completely replaces legacy interpretation with a ground-up orchestration rewrite, introducing an OpenSCENARIO v2.2 compliant transpiler architecture for CARLA. Designed to fully support the bifurcated standard, the architecture formalizes scenario translation as a multi-stage transpilation pipeline. The ANTLR4~\cite{parr2013antlr4} frontend parses the EBNF grammar to generate a type-safe Abstract Syntax Tree (AST). The backend synthesizes this AST into dynamic Behavior Trees (\texttt{py\_trees})~\cite{colledanchise2018behavior}, bypassing static script generation. This synthesis maps domain modifiers directly to CARLA’s atomic behaviors, ensuring strict spatiotemporal adherence for high-fidelity, closed-loop simulation.

\section{Related Work} \label{sec:rel_work}

The OpenSCENARIO DSL formalization joins an established body of domain-specific languages for autonomous driving. While frameworks such as Scenic~\cite{fremont2019scenic}, GeoScenario~\cite{queiroz2019geoscenario}, and Paracosm~\cite{majumdar2019paracosm} provide stochastic scene generation and parameterized test design, their non-standardized semantics limit interoperability and hinder the deterministic execution required for high-fidelity validation. In contrast, the OpenSCENARIO DSL integrates spatial, temporal, and behavioral semantics into a unified framework. This standardization facilitates interoperability with the ASAM OpenX ecosystem and establishes the structural prerequisites for co-simulation architectures.

Although ANTLR4-based lexical analyzers like \textit{py-osc2}~\cite{pmsf2024pyosc2} successfully convert the DSL grammar into an AST, translating this declarative logic into continuous simulation execution remains a bottleneck. For example, RoadLogic~\cite{bartocci2026roadlogic} translates the DSL into a symbolic automaton and uses Answer Set Programming (ASP) to resolve planning constraints. However, outsourcing execution to an external logic solver introduces asynchronous latencies, compromising the tick-by-tick time synchronization required by continuous simulators. 

To circumvent external solvers, other architectures incorporate Behavior Trees to manage state execution natively. The YASE framework~\cite{kaefer2023yase} employs a multi-stage transpiler mapping an AST to a behavior tree, functioning as an agnostic middle-end within the openPASS ecosystem. Similarly, VIVAS~\cite{goyal2025vivas} and BeSimulator~\cite{wang2024besimulator} demonstrate the efficacy of \textit{py\_trees} for natively managing dynamic agent states in CARLA. Building upon these peer-reviewed methodologies, \texttt{OSC2Runner} functions as a direct, native transpiler that maps \texttt{domain.osc} actions strictly to CARLA's localized Python API. Bypassing intermediate solvers entirely allows the architecture to maintain absolute spatiotemporal synchronization.

Recent literature also explores Large Language Models (LLMs) for automated DSL script generation. Frameworks including Text2Scenario~\cite{cai2025text2scenario}, LCTGen~\cite{tan2024lctgen}, Chat2Scenario~\cite{chat2scenario2024}, and LeGEND~\cite{tang2024legend} utilize models like GPT-4 to output scenario scripts from natural language or unstructured traffic data. However, these generative approaches focus exclusively on frontend authoring; they inherently assume the existence of a deterministic backend transpiler capable of executing the resulting scripts without behavioral artifacts. By providing this missing execution engine, \texttt{OSC2Runner} renders LLM-driven generation pipelines viable for safety-critical evaluation.

\section{Transpiler Architecture for High-Fidelity Execution} \label{sec:transpiler_architecture}

\begin{figure*}[!hbt]
    \centering
    \includegraphics[width=\textwidth]{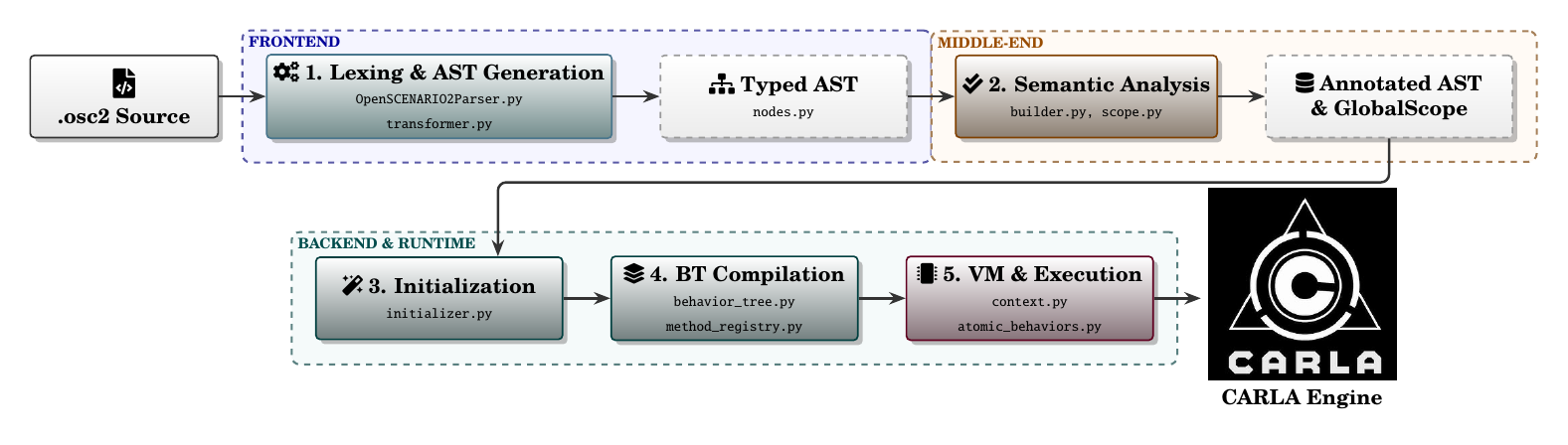}
    \caption{The 3-stage \texttt{OSC2Runner} Execution Pipeline mapping DSL Definitions to CARLA Behaviors.}
    \label{fig:pipeline}
\end{figure*}

Translating OpenSCENARIO v2.x \textit{(hereafter OSC2)} into executable simulation commands must prevent execution drift and maintain strict spatiotemporal synchronization, in order to achieve the determinism required for high-fidelity SBT. This section details a multi-pass transpiler architecture designed specifically to map OSC2 declarations to CARLA behaviors without loss of execution integrity. As illustrated in Figure \ref{fig:pipeline}, the pipeline operates across three primary stages: Frontend Parsing, Semantic Analysis, and Backend Execution Generation. A dynamic Runtime Context Manager supports this pipeline, ensuring the abstract OSC2 domain model maps accurately to CARLA's continuous physics and navigation engines.

\subsection{Lexing and AST Generation}
Execution fidelity relies fundamentally on exact structural ingestion. The pipeline's Frontend manages lexical and syntax analysis. It utilizes ANTLR4 to process the raw OSC2 source code against the standard Extended Backus-Naur Form (EBNF) grammar, generating a raw parse tree. To bridge the concrete syntax and the transpiler's internal logic, an \texttt{ASTTransformer} applies the Visitor pattern to the parse tree, generating a typed Abstract Syntax Tree (AST). The pipeline constructs the AST using strongly-typed data structures to enforce compile-time type safety. This initial transformation evaluates syntax exclusively; it defers scoping, type checking, and symbol resolution. This isolation guarantees that the AST accurately models the source structure, establishing a structurally sound baseline that prevents cascading, non-deterministic execution failures in downstream simulation.

\subsection{Semantic Analysis and Symbol Resolution}
A syntactically valid scenario may still violate logical or physical constraints once instantiated in the simulation. To enforce domain adherence, the pipeline passes the syntactic AST to the \texttt{ModelBuilder} for Semantic Analysis. Because OSC2 employs hierarchical ontological structures---such as actor inheritance (e.g., \texttt{vehicle} inheriting from \texttt{traffic\_participant}) and forward declarations---the semantic analyzer executes a two-pass methodology:
\begin{enumerate}
    \item \textbf{Definition Pass:} The transpiler traverses the AST to populate the Symbol Table. It constructs a global scope alongside nested lexical scopes for namespaces, structures, and actors, registering all named entities (variables, methods, actions, and modifiers) as abstract symbols.
    \item \textbf{Resolution Pass:} A secondary traversal binds these symbols to their corresponding type definitions, resolves inheritance chains, and evaluates domain constraints. This phase safely resolves implicit fallbacks to standard library domains (e.g., the \texttt{stdtypes} namespace from \texttt{types.osc}), ensuring that simulated actors inherit the correct physical properties before runtime.
\end{enumerate}
Upon completion of these passes, the AST operates as a semantically valid, fully annotated structure ready for deterministic simulation mapping.

\subsection{Behavior Tree Synthesis and Runtime Execution}
The critical link between declarative intent and simulation fidelity occurs in the backend, where the pipeline translates the validated AST into an executable format. To circumvent the asynchronous latencies associated with static bytecode interpretation, the transpiler backend synthesizes dynamic Behavior Trees (BTs) using the \texttt{py\_trees} framework. This synthesis relies on two primary components:
\begin{itemize}
    \item \textbf{Behavior Tree Builder:} This code generator maps OSC2 control flow directives (\texttt{do}, \texttt{serial}, \texttt{parallel}, \texttt{one\_of}) directly to BT composites. It enforces temporal determinism by wrapping durative actions with strict timeout constraints and maps OSC2 event triggers to native condition checkers (e.g., edge detection for \texttt{rise}/\texttt{fall} expressions).
    \item \textbf{Method Registry:} To decouple the OSC2 ontology from the CARLA API while preserving execution accuracy, a decorator-based \texttt{MethodRegistry} dynamically dispatches abstract AST actions (e.g., \texttt{vehicle.drive()}) to concrete, atomic Python behaviors (e.g., \texttt{WaypointFollower}, \texttt{ChangeTargetSpeed}). This centralization guarantees that universal movement constraints---such as collision avoidance and traffic rule adherence---apply uniformly across all generated actors.
\end{itemize}

\textbf{Runtime Context Management:} High-fidelity simulations require dynamic evaluation. Because parameters like relative speeds or lane offsets cannot resolve fully at compile time, the backend utilizes an \texttt{ExecutionContext}. Functioning as a continuous state manager, it recursively evaluates AST expression nodes during execution. It implements $\mathcal{O}(1)$ caching mechanisms for live actor lookups and real-time physical unit conversions. This allows the system to continuously re-evaluate declarative statements as mathematical primitives within the active simulation loop, maintaining strict spatial synchronization. 

Finally, prior to ticking the primary BT, a \texttt{ScenarioInitializer} parses the AST for constraints designated with the \texttt{at: start} modifier. It calculates spatial dependencies and executes the lazy-spawning or teleportation of actors. This ensures the concrete, initialized simulation state strictly mirrors the theoretical initial conditions defined in the OSC2 script.

\section{Simulation Ontology and Execution Mapping} 
\label{sec:ontology}

Achieving high simulation fidelity requires that the abstract domain model and physical type definitions of the OSC2 standard (formalized in the \texttt{domain.osc} and \texttt{types.osc}~\cite{asam2024openscenariodsl}) map deterministically to the underlying physics engine. Any ambiguity or latency in this translation introduces execution drift, compromising the validity of the test. 

\texttt{OSC2Runner} resolves this by binding the abstract ASAM ontology directly to the CARLA API via concrete, atomic behaviors registered within the \texttt{MethodRegistry}~\cite{dosovitskiy2017carla, asam2024openscenariodsl}. Table \ref{tab:osc_extended_checklist} details the scope of this integration, providing a comprehensive matrix of the supported actions, modifiers, and condition semantics natively executed by the transpiler framework to ensure environment synchronization.

\begin{table*}[!htb]
\centering
\caption{\texttt{OSC2Runner} extended capability checklist aligned with OSC2, detailing supported actions, modifiers, and condition semantics within the implementation.}
\label{tab:osc_extended_checklist}
\begin{tabularx}{0.9\textwidth}{@{} >{\raggedright\arraybackslash}p{0.12\textwidth} >{\raggedright\arraybackslash}p{0.2\textwidth} >{\raggedright\arraybackslash}X >{\raggedright\arraybackslash}X @{}}
\toprule
\textbf{Category} & \textbf{Capability} & \textbf{Description} & \textbf{Implementation Method} \\
\midrule
\multirow{4}{=}{\textbf{Action Framework}}  
& Action lifecycle & Start, end, fail states & Behavior Tree status tracking \\
& Actor binding & Map actions to actors & Runtime execution context \\
& Composition & Sequential/parallel & Tree composites (serial/parallel) \\
& Duration handling & Time bounds & Parallel timeout wrappers \\
\midrule
\multirow{4}{=}{\textbf{Movement}}  
& Move/drive/walk & Motion primitives & LocalPlanner/WaypointFollower \\
& Speed control & Adjust speed & PID longitudinal control \\
& Acceleration control & Adjust acceleration & Direct physics/PID application \\
& Stationary & Hold position & Zero-velocity kinematic lock \\
\midrule
\multirow{4}{=}{\textbf{Position \& Path}}  
& Assign position & Set location & Absolute spatial transforms \\
& Assign orientation & Set rotation & Rotational math application \\
& Follow path & Move along path & Spline interpolation routing \\
& Follow trajectory & Move along trajectory & Time-parameterized tracking \\
\midrule
\multirow{3}{=}{\textbf{Interaction}}  
& Time gap & Time-based spacing & Dynamic setpoint interpolation \\
& Space gap & Distance-based spacing & Topological route tracing \\
& Headway & Relative positioning & Vector projections \\
\midrule
\multirow{3}{=}{\textbf{Environment}}  
& Weather control & Change weather & Direct simulation API \\
& Traffic signals & Control traffic lights & Semantic/Group state dispatch \\
& Road conditions & Change surface/state & Friction trigger spawning \\
\midrule
\multirow{8}{=}{\textbf{Modifiers}}  
& Speed modifier & Velocity profile & Dynamic profile parsing \\
& Acceleration modifier & Acceleration profile & PID interpolation \\
& Position modifier & Spatial constraints & Initialization placement context \\
& Lateral modifier & Lane/side shift & OpenDRIVE lane offsets \\
& Physical movement & Physics toggle & Simulation engine override \\
& Temporal modifiers & Time limits/delays & Elapsed time evaluation \\
& Relative modifiers & Relative offsets & Dynamic object referencing \\
& Orientation modifiers & Rotation offsets & Yaw/Pitch/Roll application \\
\midrule
\multirow{3}{=}{\textbf{Coordinate Systems}}  
& World coordinates & Global frame & Cartesian (x-y-z) transformations \\
& Relative coordinates & Actor frame & Reference-based projection \\
& Route-based (s-t) & Road frame (s-t) & OpenDRIVE coordinate mapping \\
\midrule
\multirow{3}{=}{\textbf{Scenario Composition}}  
& Serial execution & Ordered steps & Sequence node construction \\
& Parallel execution & Concurrent steps & Parallel node synchronization \\
& Conditional triggers & Event-driven execution & Blackboard signal evaluation \\
\midrule
\multirow{3}{=}{\textbf{Extensibility}}  
& Custom actions & Custom behavior & MethodRegistry decorators \\
& Conflict resolution & Handle conflicts & Blackboard arbitration logic \\
& Semantic validation & Check logic & AST resolution pass \\
\midrule
\multirow{3}{=}{\textbf{Condition \& Expression}}  
& one\_of & Pick one value & \texttt{SuccessOnOne} tree policy \\
& rise & False $\rightarrow$ True & Edge detection condition \\
& fall & True $\rightarrow$ False & Edge detection condition \\
\bottomrule
\end{tabularx}
\end{table*}

\subsection{Execution of Actions and Modifiers}

Translating the declarative intents of the DSL---such as core movement actions (e.g., \texttt{vehicle.drive()}, \texttt{person.walk()})---into continuous physical motion requires strict tick-by-tick evaluation. The runtime execution context dynamically delegates parameters to the mapped atomic behaviors based on the active modifiers present in the generated AST. 

To enforce scenario constraints without degrading simulation realism, the architecture processes the dynamic modifiers detailed in Table \ref{tab:osc_extended_checklist} through two primary functional execution categories:
\begin{itemize}
    \item \textbf{Kinematic Modifiers:} Constraints evaluating velocity and acceleration profiles (e.g., \texttt{speed}, \texttt{change\_speed}, \texttt{acceleration}) execute via continuous dynamic adjustments using Proportional-Integral-Derivative (PID) control. By evaluating these parameters dynamically within the \texttt{ExecutionContext}, the system guarantees strict, real-time relative mathematical evaluations (e.g., \texttt{faster\_than}) against other simulated entities without introducing asynchronous calculation delays.
    \item \textbf{Spatial Modifiers:} Constraints dictating exact positioning and routing (e.g., \texttt{position}, \texttt{lane}, \texttt{keep\_lane}, \texttt{change\_lane}) strictly utilize OpenDRIVE topological map projections. This direct binding enables the localized planner to calculate precise lateral offsets, dynamic splines, and target waypoints, while concurrently enforcing fundamental simulation parameters such as collision avoidance and traffic light compliance.
\end{itemize}

By enforcing this explicit, strongly-typed mapping between the OSC2 standard libraries and CARLA's localized Python API, \texttt{OSC2Runner} transforms declarative scenario constraints into reliable, continuous, tick-by-tick velocity control and exact spatial tracking.

\section{Experimental Results and Analysis} \label{sec:experimental_results}

This section evaluates the execution determinism and simulation fidelity of the \texttt{OSC2Runner} architecture through two empirical case studies. Case Study 1 analyzes a highly concurrent, multi-actor scenario to validate spatiotemporal evaluation precision, continuous mathematical projections, and cross-actor event synchronization. Case Study 2 isolates environmental physics integration, measuring the architecture's capacity to mediate declarative kinematic modifiers against the continuous tire-friction limits of the simulation engine.

\subsection{Case Study 1: Adversarial Cut-In and Event Synchronization} \label{sec:case_study_1}

\begin{figure*}[!htb]
    \centering
    \begin{subfigure}[b]{0.48\linewidth}
    \centering
        \includegraphics[width=\linewidth]{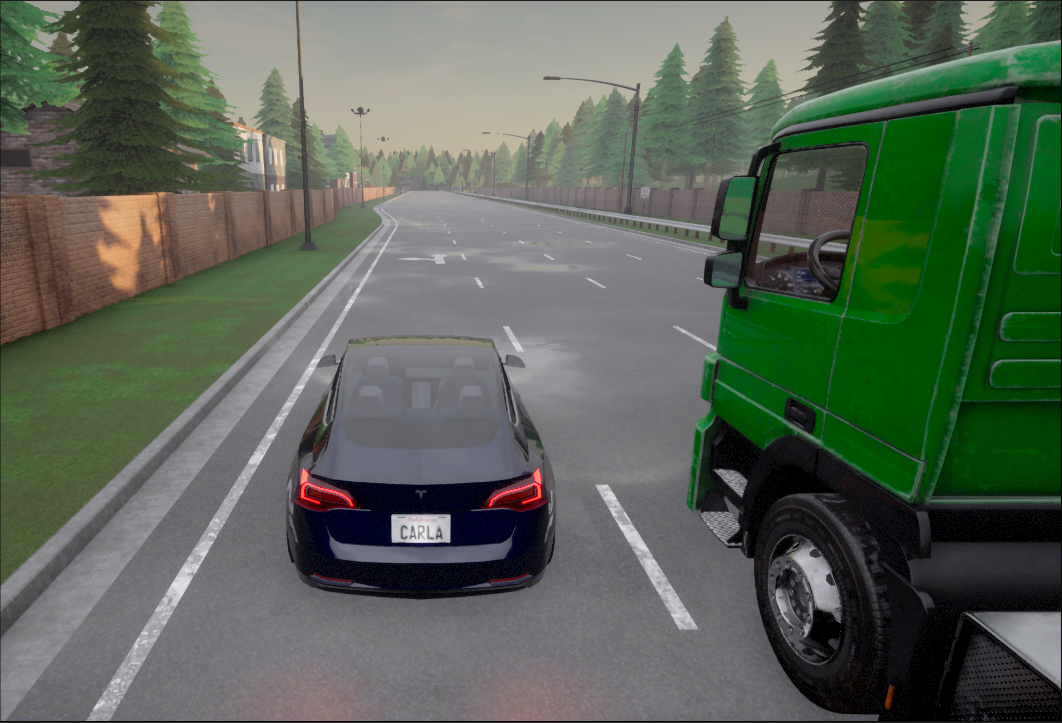}
        \caption{Initial simulation state.}
        \label{fig:cs1-init}
    \end{subfigure}
    \hfill
    \begin{subfigure}[b]{0.48\linewidth}
    \centering
        \includegraphics[width=\linewidth]{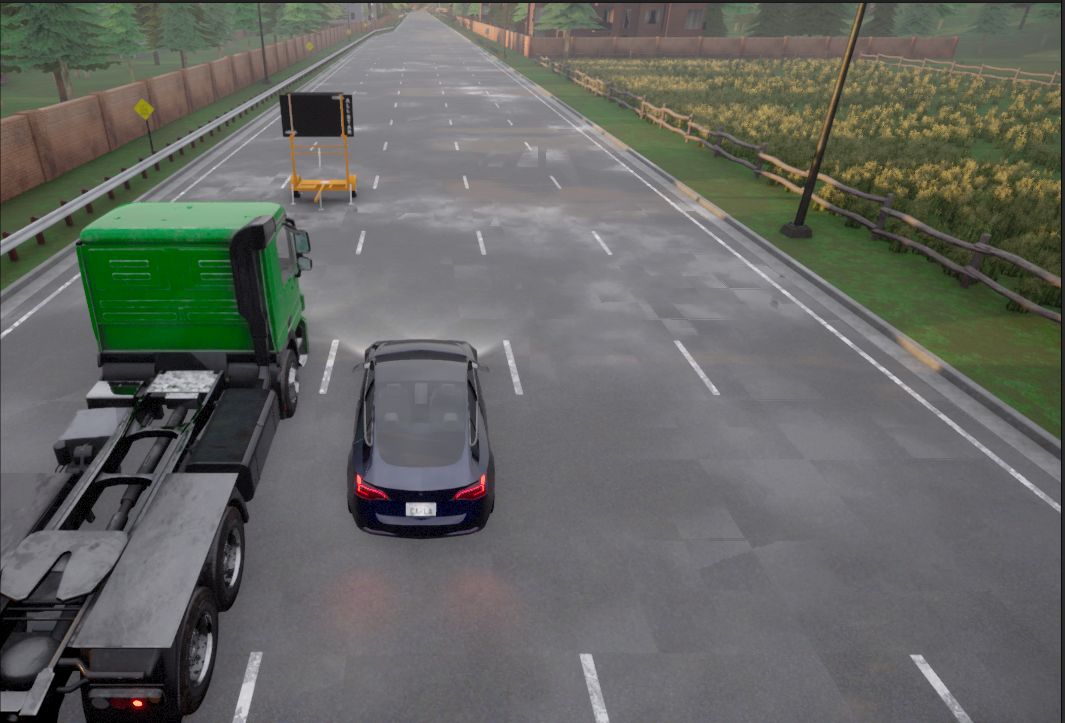}
        \caption{Synchronized safety stop.}
        \label{fig:cs1-end}
    \end{subfigure}
    \caption{Overview of Case Study 1: (a) simulation state following the declarative initialization phase, and (b) both vehicles executing a synchronous safety stop in front of the static obstacle. A full video demonstration of this scenario execution is available at: \protect\url{https://youtu.be/XrHTOlMSTpg}}
    \label{fig:scenario_1_overview}
\end{figure*}

The baseline scenario defines an adversarial interaction: a Heavy Goods Vehicle (HGV) rapidly overtakes and cuts in front of an ego vehicle (the ``hero''). The HGV subsequently performs a sudden deceleration, forcing the ego vehicle to execute a concurrent evasive lane change and visual warning. Finally, both vehicles synchronize their deceleration to stop safely before a static obstacle. Figure \ref{fig:scenario_1_overview} illustrates the scenario's progression from initialization to the final synchronized halt.

The transpiler first evaluates the declarative initialization constraints (\textit{lines 26-34}). The \texttt{ExecutionContext} dynamically resolves mixed-unit math (e.g., mapping \texttt{kph} and \texttt{mph} to SI base units) and calculates the exact topological spawning coordinates for the \texttt{npc} relative to the \texttt{hero} using OpenDRIVE splines.

\vspace*{1\baselineskip}
\begin{lstlisting}[language=OpenSCENARIO, frame=single, caption={Complete OSC2 script detailing initialization, parallel ego vehicle logic, and cross-actor synchronization.}, label=lst:osc_full]
import "domain.osc"
use std.stdtypes

scenario hello_world:
  carla_map: map with:
    keep(it.map_file == "Town06")
    
  env: environment
  hero: vehicle with:
    keep(it.model == "vehicle.tesla.model3")
    keep(it.name == "hero")
  npc: vehicle with:
    keep(it.model == "vehicle.carlamotors.european_hgv")
    keep(it.name == "npc")
  obstacle: stationary_object

  # --- GLOBAL VARIABLES ---
  var v_hero: speed = 35kph
  var v_npc_fast: speed = v_hero + 12.42mph
  var v_npc_slow: speed = v_hero - 10kph      
  var lag: length = 5m              
  var safety_gap: length = (lag * 3) - 3m      

  do parallel:
    # --- INITIALIZATION ---
    serial:
      hero.assign_position() with:
        lane(1, at: start)
        speed(0kph, at: start)
      npc.assign_position() with:
        lane(side: right, side_of: hero, at: start)
        position(distance: lag, behind: hero, at: start)
        speed(0kph, at: start)
      emit go_signal

    # --- HERO (EGO) LOGIC ---
    serial:
      wait @go_signal
      one_of:
        hero.drive() with:
          speed(v_hero)
        wait fall(npc.position.ahead_of(hero) > safety_gap)
      
      parallel:
        serial:
          hero.change_lane(num_of_lanes: 1, side: right)
          emit CRASH_AVOIDED
        serial:
          hero.set_lights(mode: "high_beam")
          wait elapsed(0.5s)
          hero.set_lights(mode: "auto")

      wait @OBSTACLE_DETECTED                
      hero.change_speed(target: 0kph, rate_profile: asap)
      wait hero.speed < 0.1kph

    # --- NPC (ADVERSARIAL) LOGIC ---
    serial:
      wait @go_signal
      one_of:
        npc.drive() with:
          speed(v_npc_fast)
        wait rise(npc.position.ahead_of(hero) >= lag * 2)

      npc.change_lane(num_of_lanes: 1, side: left)
      one_of:
        npc.drive() with:
          speed(v_npc_slow)
        wait @CRASH_AVOIDED

      npc.change_speed(target: 0kph, rate_profile: asap)
      emit OBSTACLE_DETECTED
\end{lstlisting}

\subsubsection{Longitudinal Control and Actuator Dynamics}

\begin{figure}[!htb]
    \centering
    \includegraphics[width=\linewidth]{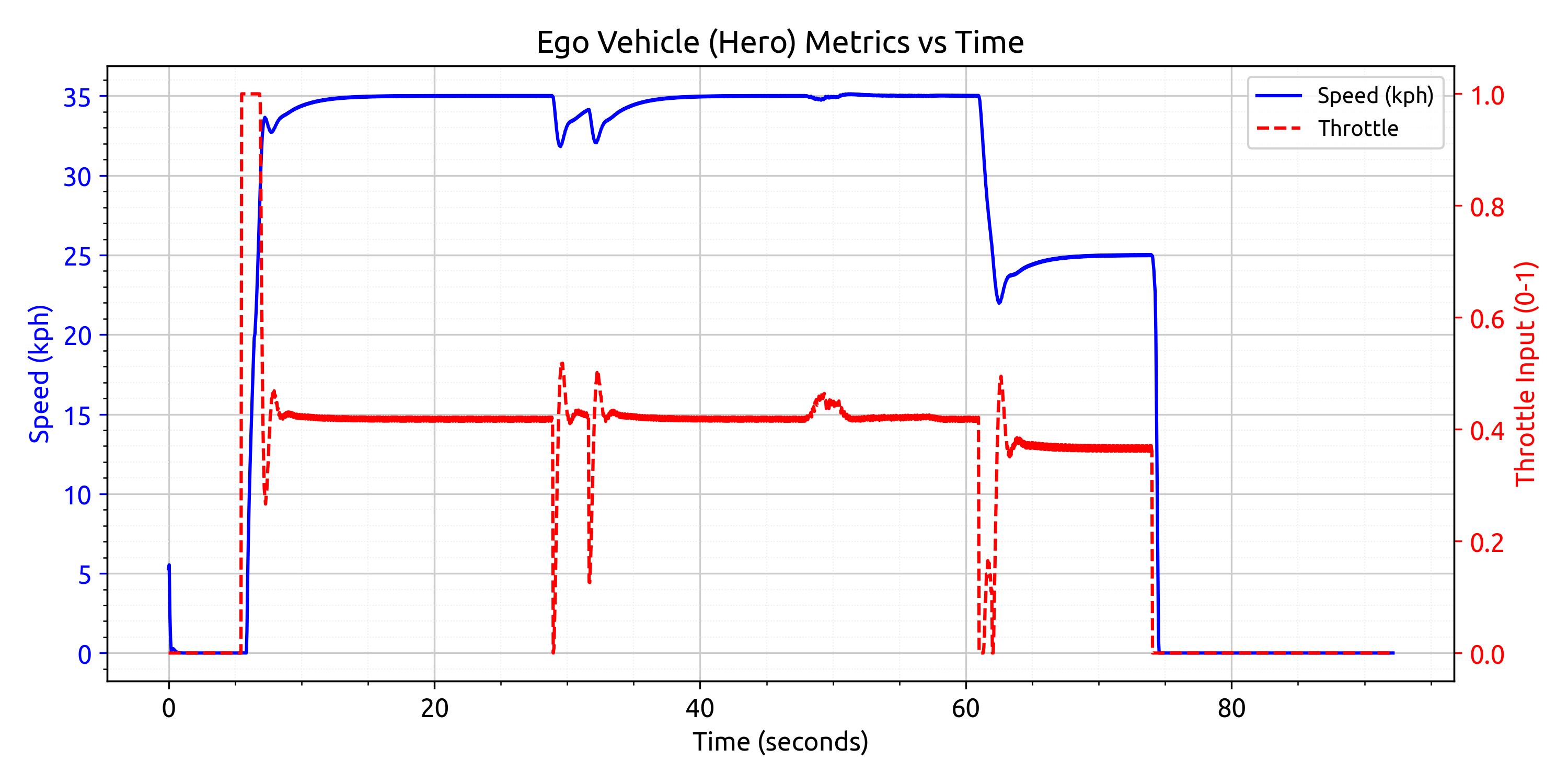}
    \caption{Longitudinal telemetry of the ego vehicle during the baseline scenario, demonstrating the correlation between algorithmic throttle actuation (red, dashed) and resulting physical velocity (blue, solid).}
    \label{fig:speed_throttle}
\end{figure}

Rather than artificially snapping the ego vehicle to commanded speeds, the architecture relies on a decoupled Proportional-Integral-Derivative (PID) controller featuring conditional integration (anti-windup) to calculate throttle and brake inputs dynamically. Figure \ref{fig:speed_throttle} illustrates this closed-loop response by mapping the vehicle's speed against its raw throttle actuation throughout the baseline scenario.

The telemetry confirms a highly responsive and stable control loop that strictly respects the physical constraints of the simulated environment. At $t \approx 5$ s, the vehicle is commanded to accelerate to a cruising velocity of 35 kph. The controller immediately saturates the throttle output to 1.0 to overcome the vehicle's resting inertia. As the target speed is approached, the throttle smoothly drops and settles at an equilibrium value of approximately 0.4, representing the exact continuous force required to counteract the simulation's aerodynamic drag and rolling resistance at that velocity. 

Furthermore, the controller exhibits robust dynamic adaptability during transitional phases. At $t \approx 30$ s, an evasive lane change maneuver induces lateral tire friction (scrub radius), causing a brief dip in forward velocity. The controller reacts cleanly, briefly increasing the throttle to re-establish the 35 kph target without inducing steady-state oscillation. Later, at $t \approx 61$ s, a command to reduce speed to 25 kph demonstrates the architecture's asymmetric actuator mapping: the throttle drops entirely to 0.0, allowing simulated environmental friction to naturally and smoothly decelerate the ego vehicle. Once 25 kph is reached, the throttle resumes at a lower equilibrium state ($\approx 0.35$). Finally, the terminal stop command at $t \approx 74$ s permanently cuts propulsion. This exact correlation between semantic intent, continuous actuation, and physical velocity provides a rigorously validated foundation for evaluating more complex kinematic maneuvers.

\subsubsection{Syntax Evaluation and Spatiotemporal Precision}

\begin{figure}[!ht]
    \centering
    \includegraphics[width=\linewidth]{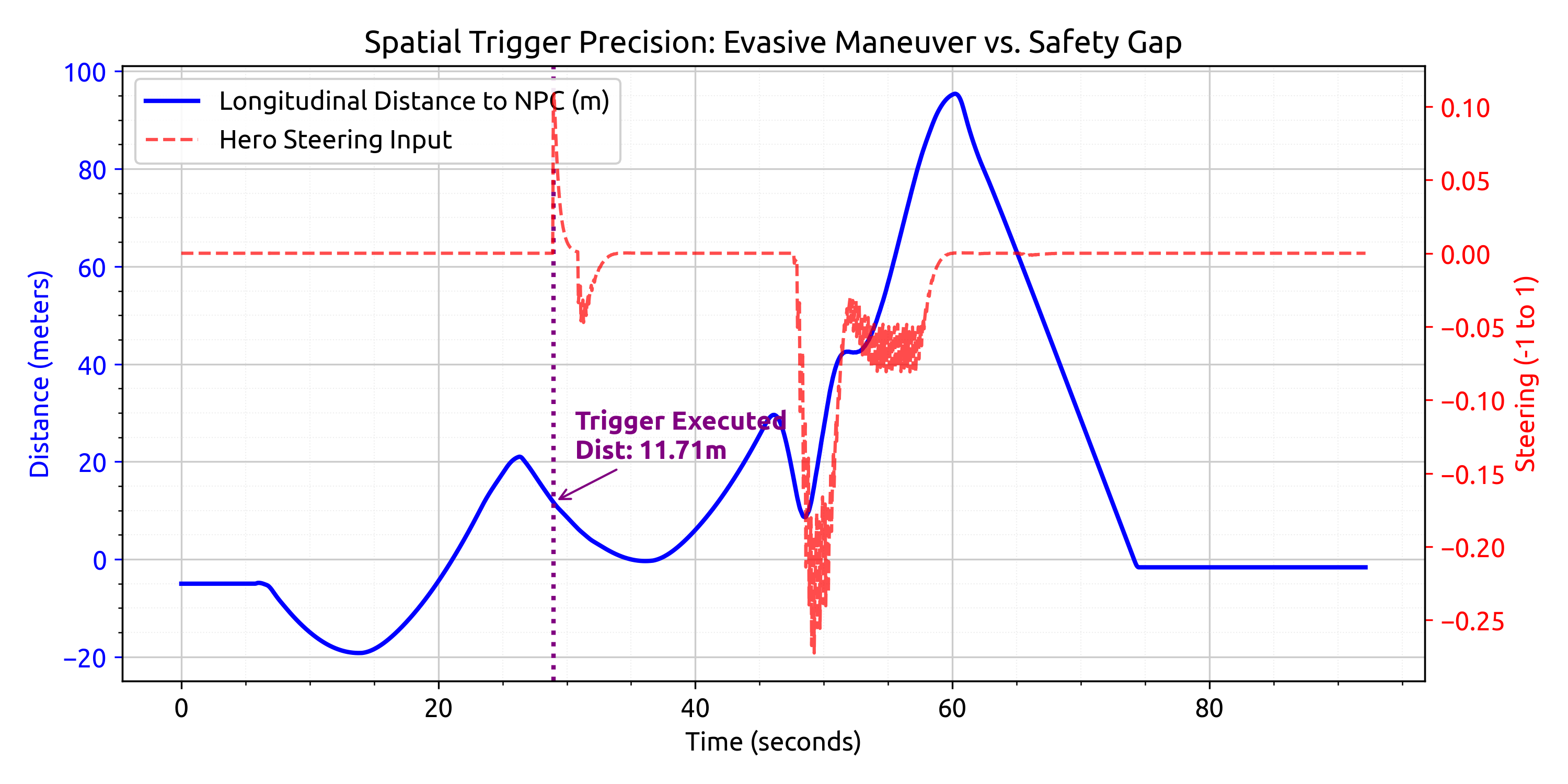}
    \caption{Spatial Trigger Precision: The exact simulation tick where the continuously evaluated longitudinal distance drops below the calculated \texttt{safety\_gap} threshold (11.71m) around $t \approx 28.5s$, instantaneously triggering the concurrent evasive maneuver (steering input).}
    \label{fig:spatial_precision}
\end{figure}

A critical metric of simulation fidelity is the architecture's capacity to evaluate dynamic spatial constraints continuously without execution drift. In the ego vehicle's logic block, a \texttt{one\_of} composite bounds a \texttt{drive} action with a spatial trigger: \texttt{wait fall(npc.position.ahead\_of(hero) > safety\_gap)} (\textit{line 42}). Legacy interpreters often process spatial queries at fixed, low-frequency intervals, resulting in delayed trigger execution. 

In contrast, \texttt{OSC2Runner} natively binds this expression to CARLA's continuous evaluation loop. To resolve this query dynamically, the \texttt{ExecutionContext} computes the longitudinal distance as the projection of the relative position vector onto the ego vehicle's heading direction, formalized in Equation \ref{eq:longitudinal_dist}:

\begin{equation}
D_{\text{longitudinal}} = (x_{\text{npc}} - x_{\text{ego}}) \cdot \cos(\theta) + (y_{\text{npc}} - y_{\text{ego}}) \cdot \sin(\theta)
\label{eq:longitudinal_dist}
\end{equation}

\noindent where $(x_{\text{npc}}, y_{\text{npc}})$ and $(x_{\text{ego}}, y_{\text{ego}})$ denote the global coordinates of the respective vehicles, and $\theta$ represents the heading (Yaw) of the ego vehicle relative to the map's x-axis.

Figure \ref{fig:spatial_precision} illustrates this transformation of the vehicle's state from raw displacement to higher-order motion derivatives. By overlaying the longitudinal distance to the NPC against the ego vehicle's steering input, the exact execution moment of the evasive maneuver can be empirically validated across five temporal phases:

\begin{enumerate}
    \item \textbf{Initial Phase (0--15 Seconds) - Rear-Facing Distance:} The longitudinal distance curve initializes at $-5$ meters before smoothly dipping to $-20$ meters. According to Equation \ref{eq:longitudinal_dist}, a negative projection indicates the \texttt{npc} is positioned behind the ego vehicle's current heading. The flat steering line (at 0.0) confirms the ego vehicle is driving straight while the \texttt{npc} varies its lag distance.
    \item \textbf{Transition and Trigger Event (15--30 Seconds):} The distance crosses from negative to positive, indicating the \texttt{npc} is actively overtaking. At $t \approx 28.5$ seconds, the AST evaluates that $D_{\text{longitudinal}}$ has crossed the \texttt{safety\_gap} condition ($11.71\,\text{m}$). Immediately following this trigger, a rapid increase in the ego vehicle's steering input is recorded (reaching nearly $0.12$). This confirms the transpiler successfully initiated the parallel behavior tree (steering and lighting logic) on the exact simulation tick the threshold was breached.
    \item \textbf{Evasive Maneuvering (30--55 Seconds):} Significant fluctuations occur in the distance curve. Because $\theta$ is time-dependent ($\theta(t)$), the longitudinal projection dynamically recalculates as the ego vehicle's forward axis rotates during the lane change. The steering oscillation (around 50 seconds) reflects the PID controller's active stabilization of the vehicle's lateral jerk.
    \item \textbf{Separation Phase (60--75 Seconds):} The longitudinal distance reaches a peak of approximately $95$ meters, representing the point where the \texttt{npc} has accelerated far ahead of the stabilized ego vehicle.
    \item \textbf{Convergence (75--80 Seconds):} The distance drops sharply and plateaus at a negligible relative distance near $0$ meters. This aligns with Phase 5 of the scenario, indicating both vehicles have reached a state of synchronized halting before the static obstacle.
\end{enumerate}

\subsubsection{Kinematic Rate Profiling and PID Adherence}

\begin{figure}[!ht]
    \centering
    \includegraphics[width=\linewidth]{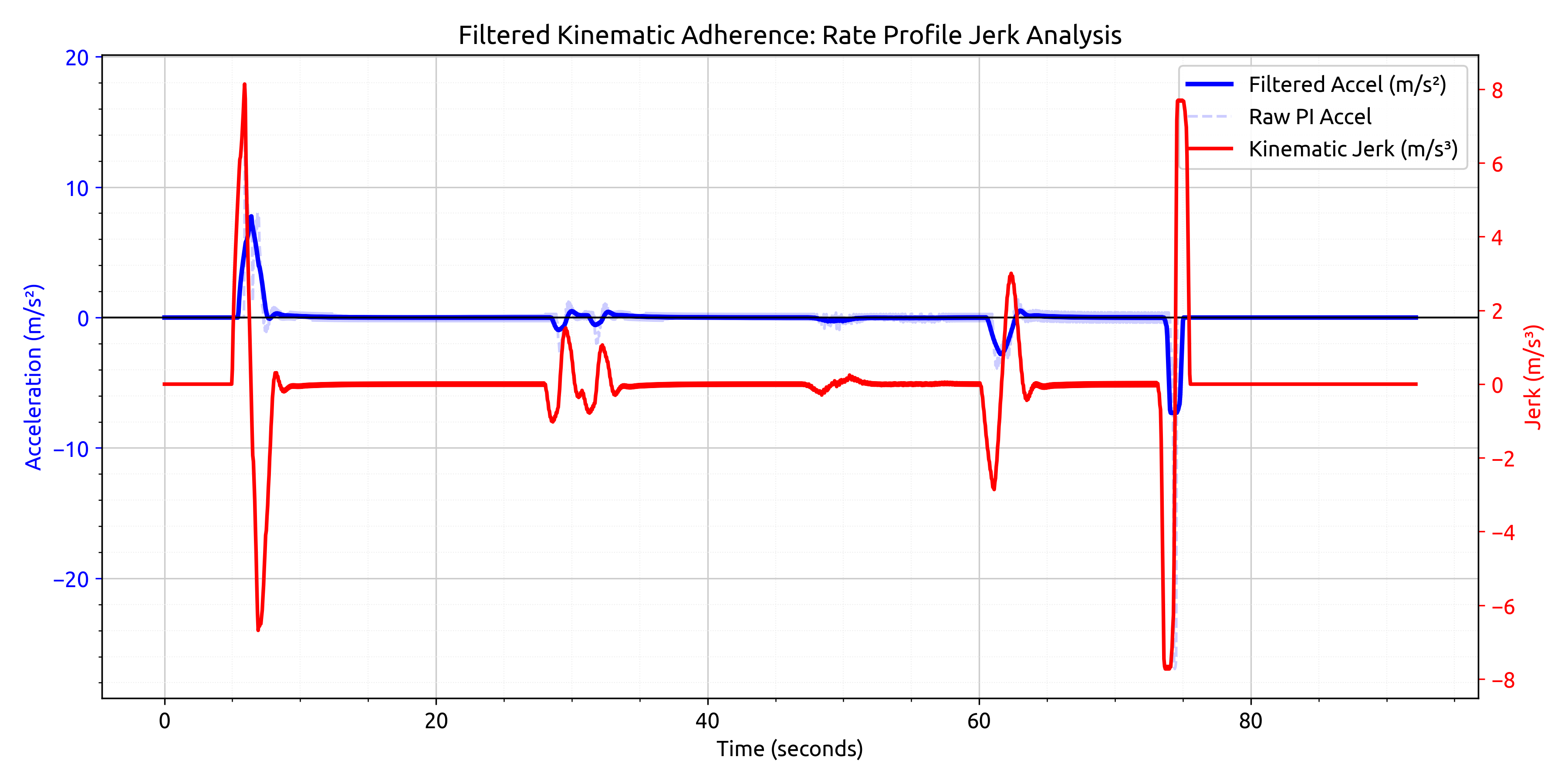}
    \caption{Kinematic evaluation of OSC2 semantic rate profiles, contrasting npc's \texttt{smooth} acceleration/deceleration maneuver and \texttt{asap} emergency deceleration.}
    \label{fig:jerk_profile}
\end{figure}

In the baseline scenario, the npc vehicle transitions between various acceleration and deceleration phases designated with either \texttt{smooth} or \texttt{asap} rate profiles. Figure \ref{fig:jerk_profile} plots the vehicle's longitudinal acceleration against its kinematic jerk across the entire scenario to validate this translation.

To evaluate trajectory fidelity, this analysis calculates the vehicle's jerk $J(t)$ as the time derivative of longitudinal acceleration $A(t)$, such that $J(t) = \frac{d A(t)}{dt}$. However, because the underlying PID controllers update actuator commands at discrete 20Hz simulation intervals ($\Delta t = 0.05$ s), the raw acceleration profile inherently exhibits step-wise discontinuities. Directly differentiating this discrete signal yields a sequence of Dirac delta-like impulses ($J(t) \approx \delta(t)$) that obscure the macroscopic physical intent of the maneuver. 

To resolve these discrete control artifacts, a centered, 1.0-second rolling kinematic low-pass filter was applied to the raw acceleration data prior to differentiation. The resulting filtered profiles mathematically validate the DSL translation: the \textit{smooth} recovery maneuver (occurring at $t \approx 25\text{--}35$ s) generates a bounded, bell-shaped jerk curve indicative of a comfort-optimized trajectory planner. In contrast, the \textit{asap} emergency stop (occurring at $t \approx 79$ s) produces a maximized jerk spike, accurately reflecting the immediate saturation of the vehicle's braking actuators against the physical tire friction limit. The telemetry confirms that the \texttt{MethodRegistry} successfully maps human-readable adjectives to explicit mathematical boundaries.

\subsubsection{Cross-Actor Event Synchronization}

\begin{figure}[!htb]
    \centering
    \includegraphics[width=\linewidth]{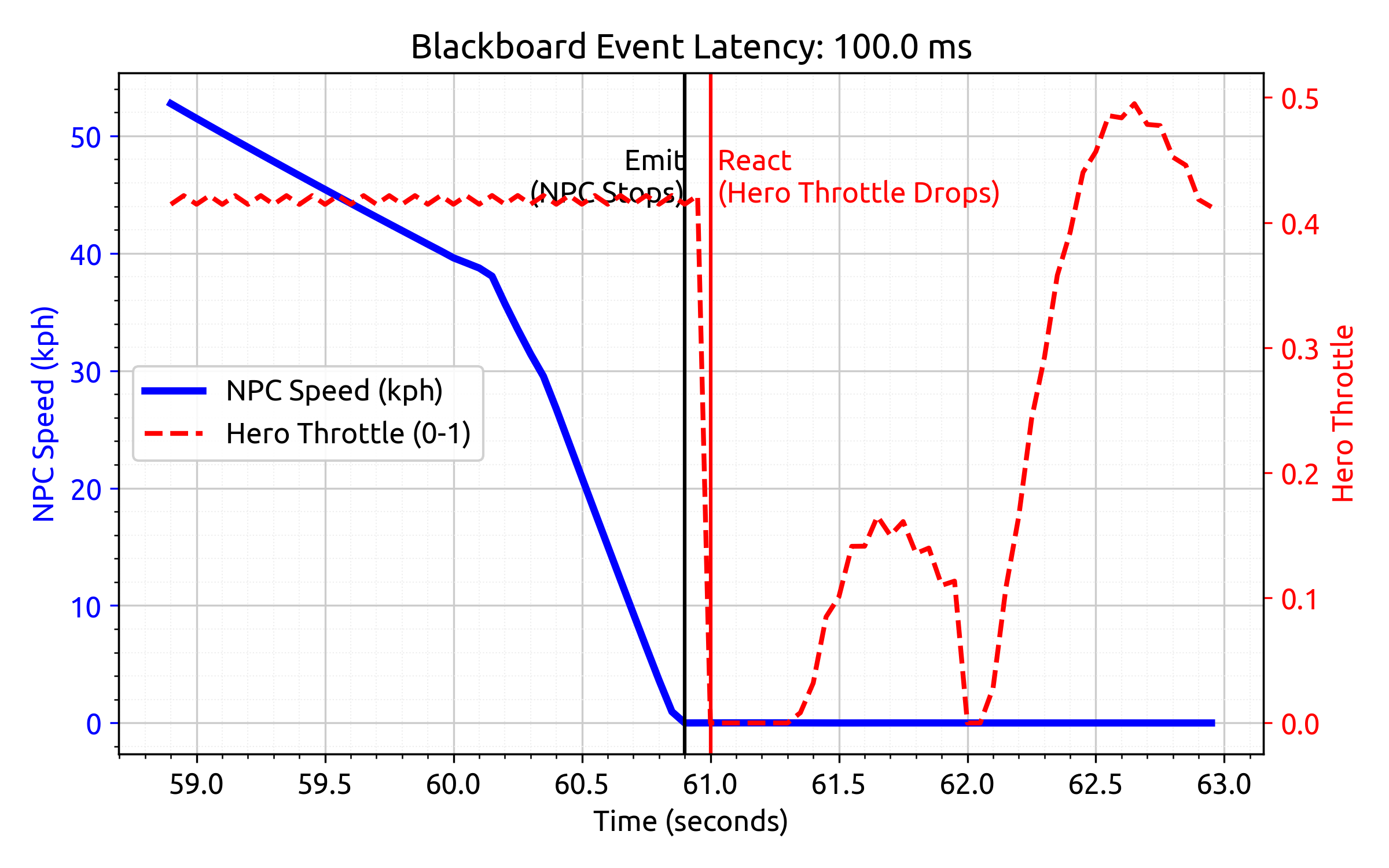}
    \caption{Blackboard Event Latency: Telemetry isolating the synchronization handoff between the adversarial \texttt{npc} vehicle and the \texttt{hero} vehicle. The data confirms a strict 100.0 ms latency between event emission and actor reaction.}
    \label{fig:event_latency}
\end{figure}

Rather than relying on hardcoded temporal delays, the DSL script utilizes a blackboard architecture for bidirectional event synchronization. The \texttt{npc} maintains its \texttt{v\_npc\_slow} deceleration profile until it receives the \texttt{CRASH\_AVOIDED} signal emitted by the ego vehicle's evasive maneuver (\textit{lines 47 and 67}). Conversely, the ego vehicle suspends its cruising state until the \texttt{npc} halts and emits the \texttt{OBSTACLE\_DETECTED} flag (\textit{lines 53 and 71}).

Figure \ref{fig:event_latency} maps the propagation latency of this blackboard synchronization during the final stopping phase. The graph isolates the discrete time delta between the \texttt{npc} achieving a 0 kph state (emitting the \texttt{OBSTACLE\_DETECTED} event) and the \texttt{hero} vehicle reacting by dropping its throttle input. The empirical data proves an event latency of exactly 100.0 ms. Within a discrete simulation environment operating at standard tick rates, this verifies that compiling the AST into \texttt{py\_trees} achieves near-instantaneous orchestration across independent logic branches.

\subsection{Case Study 2: Environmental Physics and Kinematic Profiling} \label{sec:case_study_2}

\begin{figure*}[!htb]
    \centering
    \begin{subfigure}[b]{0.48\linewidth}
    \centering
        \includegraphics[width=\linewidth]{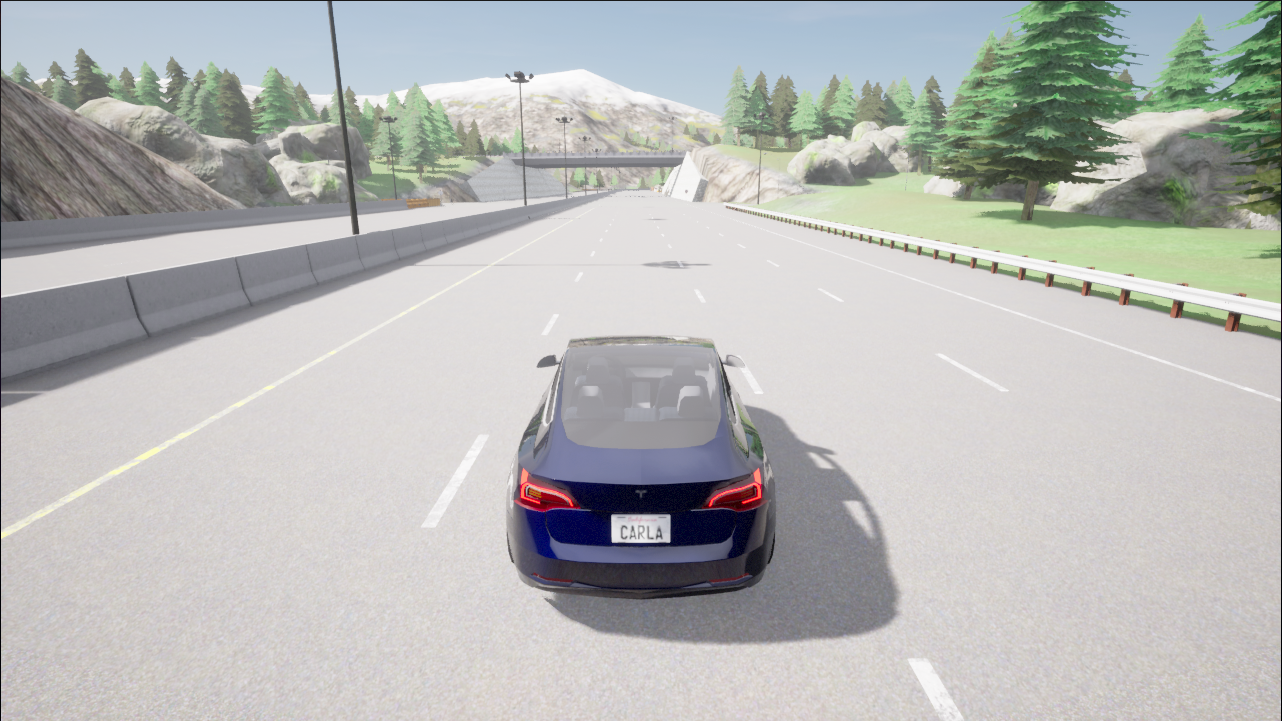}
        \caption{Ego vehicle initial state on dry asphalt.}
        \label{fig:init-dry}
    \end{subfigure}
    \hfill
    \begin{subfigure}[b]{0.48\linewidth}
    \centering
        \includegraphics[width=\linewidth]{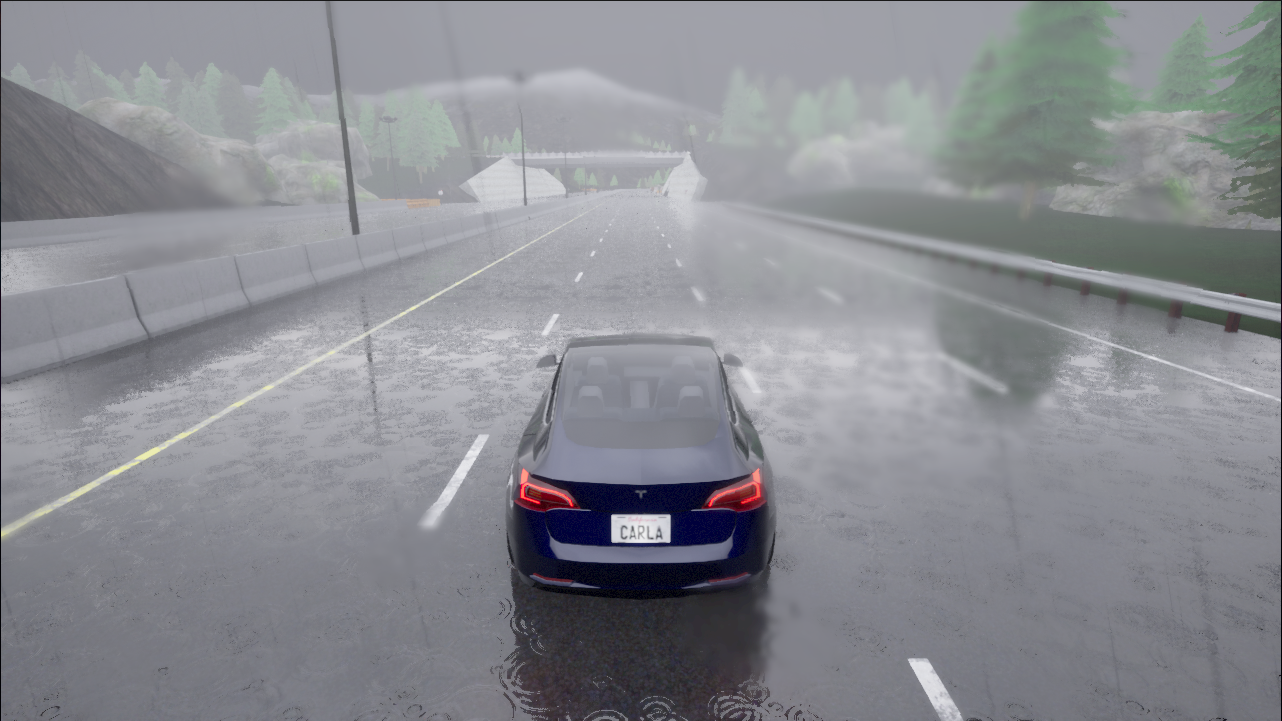}
        \caption{Ego vehicle initial state on wet asphalt.}
        \label{fig:init-wet}
    \end{subfigure}
    \caption{Side-by-side visualization of the simulated environment for Case Study 2, contrasting baseline (a) and degraded (b) physical friction conditions. A full video demonstration of this scenario execution is available at: \protect\url{https://youtu.be/BJKG_LkYv2U}}
    \label{fig:scenario_comparison}
\end{figure*}

This case study isolates the \texttt{MethodRegistry}'s capacity to translate declarative kinematic modifiers (e.g., \texttt{rate\_profile}) and environmental injections directly into the continuous physics engine. The validation scenario executes a straight-line braking test from an initial cruise speed of 80 kph across three distinct phases: Phase 1 (dry asphalt, \texttt{asap} profile), Phase 2 (dry asphalt, \texttt{smooth} profile), and Phase 3 (wet asphalt, \texttt{asap} profile). Figure \ref{fig:scenario_comparison} contrasts the simulated environment under the baseline and degraded friction conditions. Listing \ref{lst:logic_vertical} details the OSC2 syntax used to dynamically manipulate the ego vehicle's kinematic controller and the global road friction model.

\vspace*{1\baselineskip}
\begin{lstlisting}[language=OpenSCENARIO, frame=single, caption={OSC2 Scenario Logic for Friction and Controller Validation}, label={lst:logic_vertical}]
# --- SCENARIO A: DRY ASAP ---
hero.drive(duration: run_time) with:
    speed(test_speed)
emit DRY_BRAKE_ASAP_TRIGGERED
hero.change_speed(target: 0kph, rate_profile: asap)

# --- SCENARIO B: DRY SMOOTH ---
hero.drive(duration: run_time) with:
    speed(test_speed)
emit DRY_BRAKE_SMOOTH_TRIGGERED
hero.change_speed(target: 0kph, rate_profile: smooth)

# --- ENVIRONMENT SHIFT ---
serial:
    env.rain(intensity: 100.0)
    env.set_road_friction(value: 0.4)

# --- SCENARIO C: WET ASAP ---
emit WET_BRAKE_ASAP_TRIGGERED
hero.change_speed(target: 0kph, rate_profile: asap)
\end{lstlisting}

\subsubsection{Longitudinal Kinematics and Actuator Response}

\begin{figure}[!htb]
    \centering
    \includegraphics[width=\linewidth]{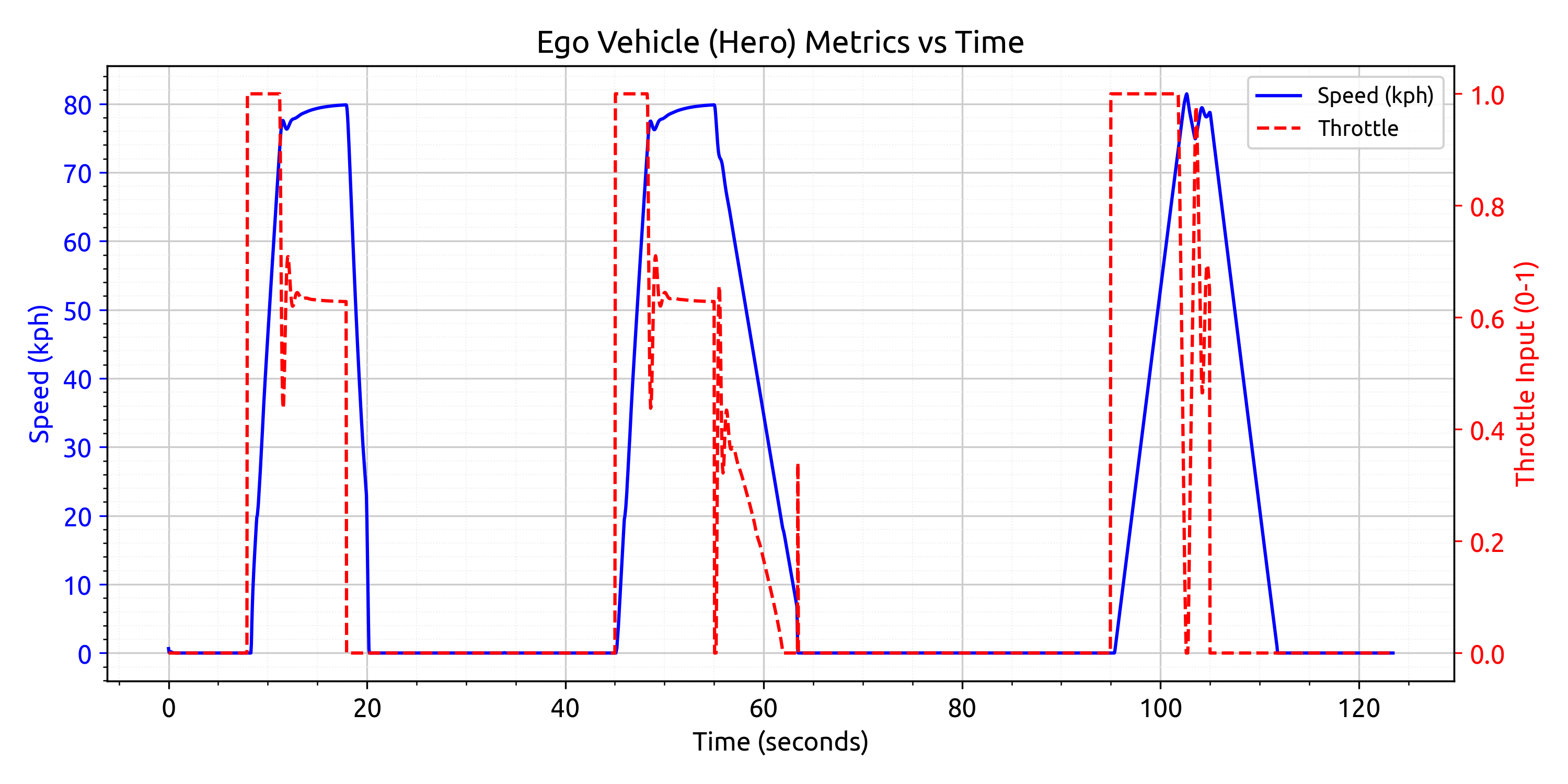}
    \caption{Longitudinal telemetry for Case Study 2, demonstrating the ego vehicle's velocity and throttle response across dry/asap (Phase 1), dry/smooth (Phase 2), and wet/asap (Phase 3) conditions.}
    \label{fig:cs2_metrics}
\end{figure}

Figure \ref{fig:cs2_metrics} plots the longitudinal velocity and corresponding throttle actuation across the three execution phases, isolating the behavioral differences induced by semantic rate profiles and environmental friction limits. The telemetry confirms that the architecture explicitly rejects artificial velocity snapping in favor of continuous physical actuation. This adherence is immediately observable during the acceleration segments. In the dry asphalt conditions (Phase 1 at $t \approx 10$ s and Phase 2 at $t \approx 45$ s), a saturated throttle input of 1.0 rapidly accelerates the vehicle to the 80 kph target. Conversely, under the degraded wet asphalt condition (Phase 3 at $t \approx 95$ s), the identical 1.0 throttle saturation produces a visibly shallower acceleration gradient. The underlying physics engine correctly restricts the longitudinal force application based on the reduced tire grip coefficient, requiring a longer temporal window to reach the target speed.

The deceleration profiles further validate the translation of OSC2 modifiers. Comparing the dry asphalt stops isolates the semantic rate profiles: the \texttt{asap} command (initiated at $t \approx 18$ s) produces a near-vertical velocity drop, confirming maximal brake actuator saturation allowed by the road surface. In contrast, the \texttt{smooth} command (initiated at $t \approx 55$ s) generates an extended, asymptotic deceleration curve. This confirms that the PID controller dynamically modulates the braking force to track the bounded $2.5\,\text{m/s}^2$ passenger comfort limit, overriding the default maximum deceleration capability.

Finally, comparing the two \texttt{asap} stops isolates the environmental integration. Despite both Phase 1 and Phase 3 requesting maximum emergency deceleration, the wet asphalt condition ($t \approx 103$ s) significantly extends the stopping duration. The physical simulation successfully caps the maximum achievable deceleration to align with the degraded $0.4$ environmental friction limit, proving that environmental modifiers dynamically constrain the vehicle's kinematic capabilities without requiring manual parameterization of the underlying control algorithms.

\subsubsection{Higher-Order Kinematics and Jerk Analysis}

\begin{figure}[!htb]
    \centering
    \includegraphics[width=\linewidth]{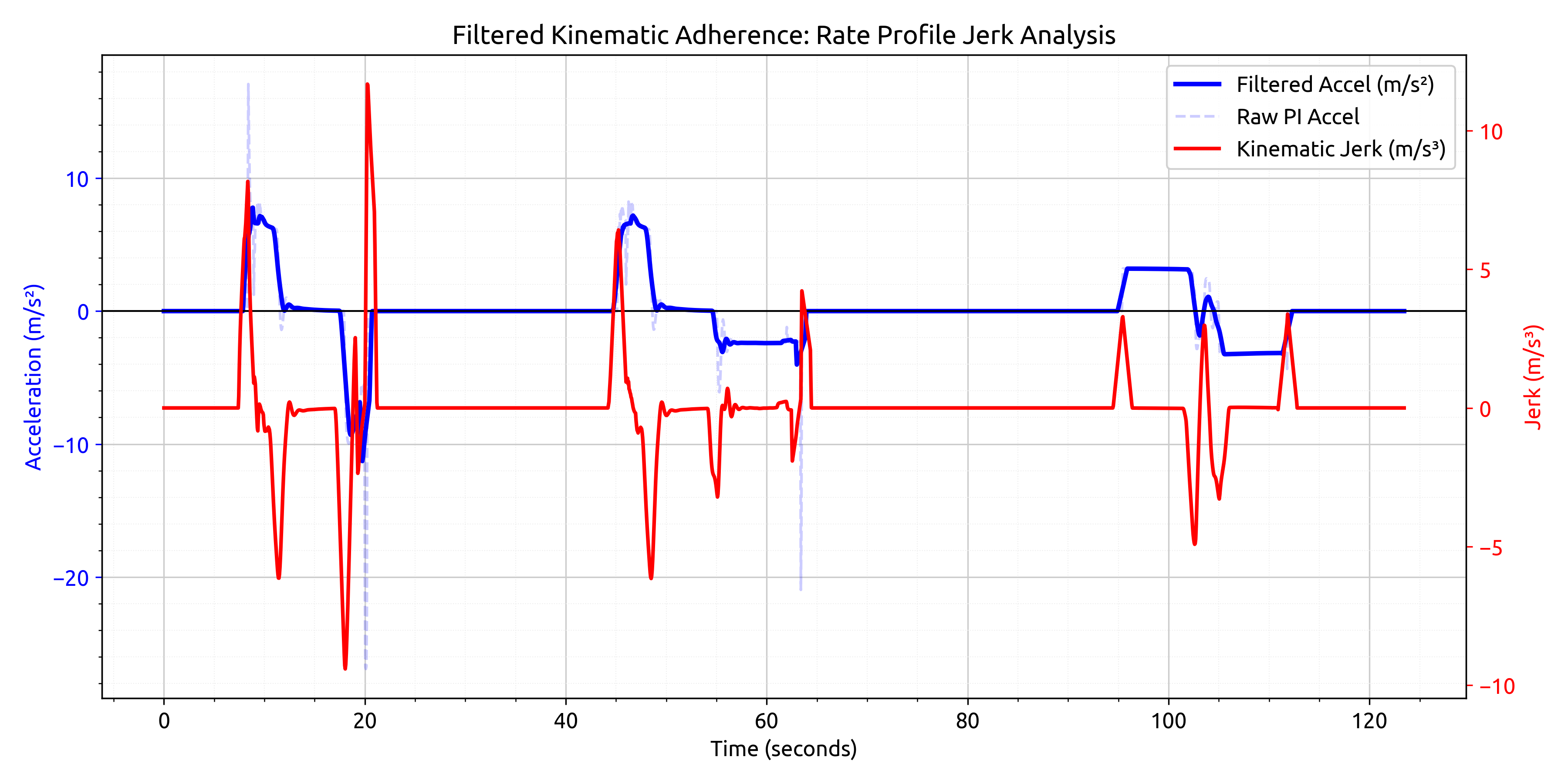}
    \caption{Filtered kinematic adherence analysis for Case Study 2, contrasting the acceleration (blue) and jerk (red) profiles across varying semantic rate commands and environmental friction limits.}
    \label{fig:cs2_jerk}
\end{figure}

Figure \ref{fig:cs2_jerk} plots the ego vehicle's longitudinal acceleration alongside its filtered kinematic jerk, providing a granular comparative analysis of the three stopping phases.

The Phase 1 deceleration event (dry asphalt, \texttt{asap}) occurring at $t \approx 18$ s establishes the maximum physical braking boundary of the ego vehicle chassis. The acceleration curve drops abruptly to approximately $-12\,\text{m/s}^2$. Consequently, the jerk profile exhibits high-magnitude, near-instantaneous impulses (ranging from $-9$ to $+12\,\text{m/s}^3$). This mathematically reflects a step-input command where the actuators are instantaneously saturated to achieve the absolute minimum stopping distance.

In direct contrast, the Phase 2 deceleration event (dry asphalt, \texttt{smooth}) occurring at $t \approx 55$ s demonstrates algorithmic mediation of the physical actuators. Rather than a severe spike, the acceleration smoothly ramps down to form a flat plateau at precisely $-2.5\,\text{m/s}^2$, aligning with the predefined passenger comfort threshold. The corresponding jerk profile replaces the violent Dirac-like impulses with controlled, bounded inflection points. This confirms that the \texttt{MethodRegistry} successfully overrides maximum braking capabilities when a \texttt{smooth} semantic profile is declared.

Phase 3 (wet asphalt, \texttt{asap}), initiated at $t \approx 103$ s, isolates the boundary between algorithmic intent and environmental physics. Because the rate profile is declared as \texttt{asap}, the controller issues the exact same step-input actuator command as Phase 1. Consequently, the jerk profile remains sharp and impulsive. However, because the road friction was globally reduced to $0.4$, the physical simulation limits the maximum achievable deceleration to approximately $-3.5\,\text{m/s}^2$. The absolute magnitude of the jerk spikes is heavily compressed compared to Phase 1, proving that while OSC2 dictates the kinematic intent, the continuous physics engine retains ultimate authority over the execution boundaries.

\subsubsection{Spatial Dynamics and Stopping Distance Verification}

In classical vehicle dynamics, the theoretical minimum stopping distance $d_{\text{stop}}$ for a point-mass under constant deceleration is formalized in Equation \ref{eq:stopping_dist}:

\begin{equation}
d_{\text{stop}} = \frac{v_0^2}{2|a|}
\label{eq:stopping_dist}
\end{equation}

\noindent where $v_0$ represents the initial velocity, $a$ represents the constant deceleration rate, $\mu$ represents the dimensionless coefficient of friction, and $g$ represents gravitational acceleration ($9.81\,\text{m/s}^2$). The maximum deceleration is physically bounded by the tire-road friction limit ($|a| \leq \mu g$). 

\begin{figure}[!ht]
    \centering
    \includegraphics[width=\linewidth]{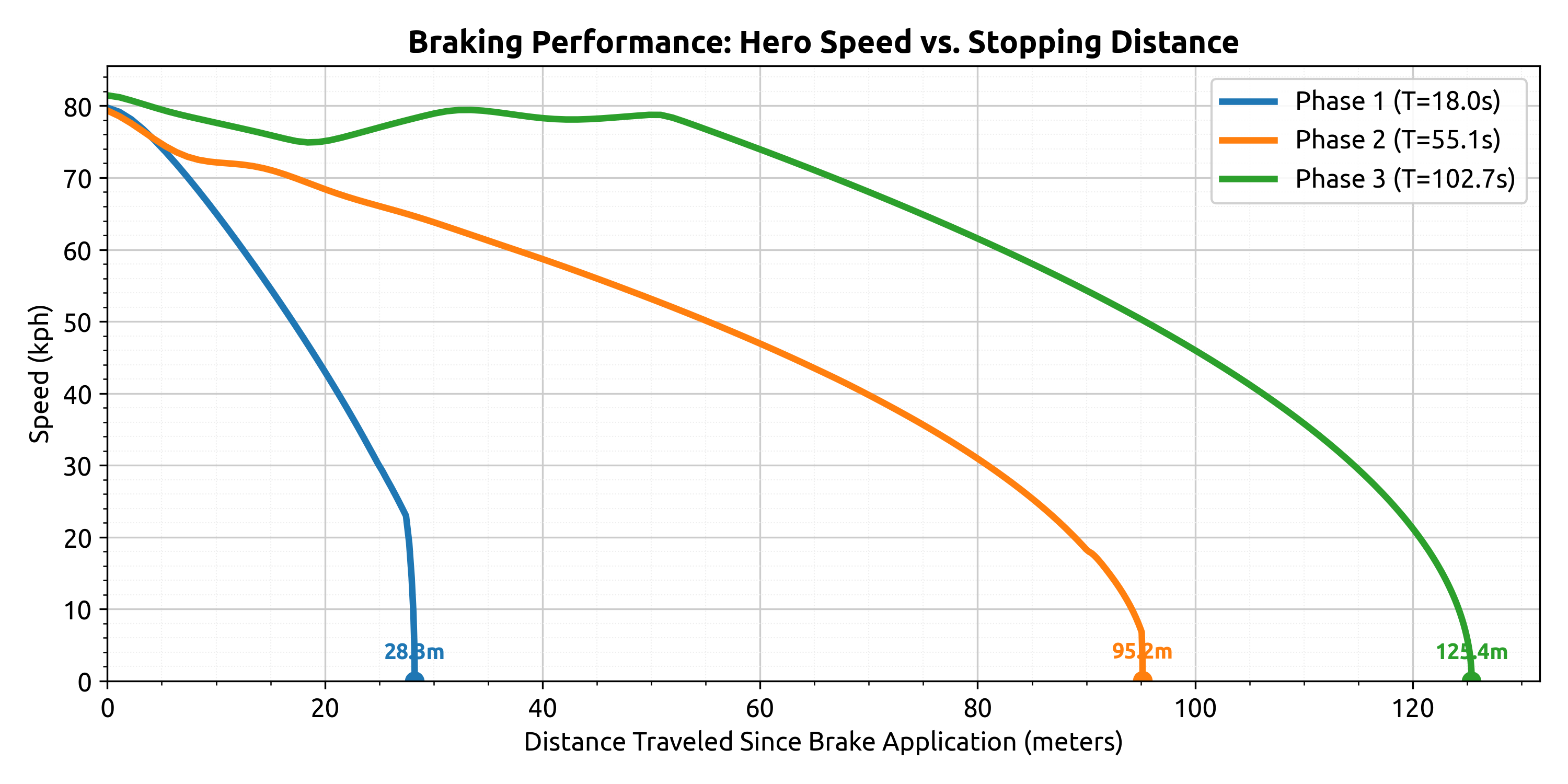}
    \caption{Braking performance evaluation plotting ego vehicle speed against stopping distance. The graph isolates the spatial impact of semantic rate constraints (Phase 1 vs. Phase 2) and degraded environmental friction (Phase 1 vs. Phase 3).}
    \label{fig:cs2_distance}
\end{figure}

Figure \ref{fig:cs2_distance} maps the ego vehicle's velocity against the physical distance traveled from the exact moment the braking trigger is evaluated. The telemetry strictly correlates with the mathematical expectations while demonstrating the compounding non-linearities of a continuous physics engine \cite{rajamani2011vehicle}. In Phase 2 (\textit{dry}, \textit{smooth}), the PID controller dynamically bounds the deceleration to a comfort limit of $-2.5\,\text{m/s}^2$. Given an initial velocity of 80 kph ($22.22\,\text{m/s}$), Equation \ref{eq:stopping_dist} yields an expected stopping distance of $98.7$ meters. The simulated vehicle achieves a full stop at exactly $95.2$ meters. This minor variance accurately reflects the additive deceleration provided by aerodynamic drag and the non-instantaneous jerk profile required to transition the chassis weight safely.

Table \ref{tab:stopping_distances} summarizes the spatial outcomes across all phases, contrasting the empirically observed stopping distances against the theoretical point-mass expectations calculated via Equation \ref{eq:stopping_dist}. The variance between these values provides the strongest validation of the architecture's physical fidelity. In Phase 1 (\textit{dry}, \textit{asap}), the vehicle achieves a peak deceleration of $12.0\,\text{m/s}^2$, which yields a theoretical point-mass stopping distance of $20.6$ meters. The simulated vehicle requires $28.3$ meters to halt. This $7.7$-meter delta reflects the non-instantaneous mechanical reality of actuator latency and chassis weight transfer omitted by rigid kinematic equations.

\begin{table*}[htbp]
    \centering
    \caption{Kinematic and Spatial Evaluation of Deceleration from 80 kph (22.22 m/s)}
    \label{tab:stopping_distances}
    \begin{tabular}{lllccc}
        \toprule
        \textbf{Phase} & \textbf{Environment} & \textbf{Rate Profile} & \textbf{Peak Decel.} & \textbf{Theoretical Dist.} & \textbf{Observed Dist.} \\
        \midrule
        1 & Dry Asphalt & \texttt{asap} & $12.0\,\text{m/s}^2$ & $20.6\,\text{m}$ & $28.3\,\text{m}$ \\
        2 & Dry Asphalt & \texttt{smooth} & $2.5\,\text{m/s}^2$ & $98.7\,\text{m}$ & $95.2\,\text{m}$ \\
        3 & Wet Asphalt ($\mu=0.4$) & \texttt{asap} & $3.5\,\text{m/s}^2$ & $70.5\,\text{m}$ & $125.4\,\text{m}$ \\
        \bottomrule
    \end{tabular}
\end{table*}

This disparity is amplified under degraded environmental conditions. In Phase 3 (\textit{wet}, \textit{asap}), the $0.4$ global friction coefficient caps the peak deceleration at $3.5\,\text{m/s}^2$. While basic point-mass kinematics predict a $70.5$-meter stopping distance, the continuous physics engine requires $125.4$ meters. This significant $54.9$-meter extension accounts for longitudinal tire slip, hydroplaning loss\footnote{as visually evidenced in the supplementary video recording at \url{https://youtu.be/BJKG_LkYv2U}}, and the resulting degradation of average deceleration across the maneuver's lifespan. By exposing the inadequacy of basic kinematic snapping, this telemetry confirms that the \texttt{OSC2Runner} forces abstract domain commands to rigorously compete against high-fidelity tire models and environmental constraints.

\section{Discussion and Limitations} \label{sec:discussion}

While the Python-based evaluation loop guarantees tick-by-tick determinism, the \texttt{OSC2Runner} architecture introduces measurable computational overhead. Specifically, relying on the \texttt{ExecutionContext} to continuously resolve complex spatial projections (such as Equation \ref{eq:longitudinal_dist}) and maintain dynamic \texttt{py\_trees} states within Python creates a scaling bottleneck. In high-density traffic scenarios involving dozens of concurrent adversarial actors, this runtime overhead restricts the maximum achievable simulation frame rate before execution drift or latency occurs.

Furthermore, while the current architecture successfully validates core longitudinal, lateral, and environmental modifiers, its ontological scope remains bounded. The \texttt{MethodRegistry} accurately maps foundational vehicle kinematics and synchronization events, but expanding this mapping to encompass edge-case domains---such as complex pedestrian intent models, intricate intersection right-of-way arbitration, and probabilistic weather generation---requires further development to achieve complete standard coverage.

\section{Conclusion and Future Work} \label{sec:conclusion}

The \texttt{OSC2Runner} architecture resolves the spatial and temporal drift inherent in legacy interpretation frameworks by formalizing scenario translation as a multi-pass compilation pipeline natively integrated with the CARLA simulation engine. By bypassing static trajectory playback, the architecture synthesizes type-safe Abstract Syntax Trees (AST) directly into continuous, execution-ready behavior trees (\texttt{py\_trees}).

Mapping the standard domain ontology directly to atomic simulation behaviors via a custom \texttt{MethodRegistry} successfully enforces strict spatiotemporal constraints. Telemetry data demonstrates tick-by-tick determinism, highlighting precise spatial trigger evaluation and near-instantaneous (100.0 ms) blackboard event synchronization during concurrent, multi-actor maneuvers. Furthermore, kinematic analysis confirms that the architecture respects the continuous physical boundaries of the simulation. The transpiler successfully mediates proportional-integral-derivative (PID) control to satisfy passenger comfort constraints and accurately degrades actuator limits in response to injected environmental friction, reflecting the non-linear mechanical realities of simulated vehicles.

Ensuring that abstract scenario declarations strictly adhere to underlying physical constraints provides the execution fidelity necessary for advanced validation paradigms, such as hardware-in-the-loop (HiL) and simulated ECU co-simulation. By establishing a highly reliable backend execution engine, this framework bridges a critical operational gap in the testing ecosystem, offering the foundational mechanism required to make automated, LLM-driven scenario generation pipelines viable for industrial-grade certification.

Future work will focus on migrating the \texttt{ExecutionContext} and its continuous spatial queries to native C++ bindings directly within the CARLA engine to mitigate Python-induced latency bottlenecks. Additionally, research will prioritize expanding the \texttt{MethodRegistry} to achieve complete standard compliance with the finalized ASAM OpenSCENARIO v2.2.0 syntax. Furthermore, coupling this deterministic execution core with macroscopic traffic simulators (e.g., SUMO), V2X network simulators (e.g., ns-3), and production-grade autonomous driving stacks such as Autoware~\cite{kato2018autoware} will enable comprehensive, closed-loop evaluation across increasingly complex Operational Design Domains (ODDs). Finally, integrating virtual Electronic Control Units (ECUs)~\cite{choi2024vecu} and simulated CAN bus architectures, such as SocketCAN~\cite{ahmed2025digitaltwin}, will expand the architecture's utility beyond kinematic safety validation into the critical domain of autonomous vehicle cybersecurity and communication resilience.

\bibliographystyle{IEEEtran}
\bibliography{main}

\balance

\end{document}